\documentclass[11pt,a4paper]{article}
\usepackage[hyperref]{emnlp2020}
\usepackage{times}
\usepackage{latexsym}
\usepackage{amsmath}
\usepackage{mathrsfs}
\usepackage{booktabs}
\usepackage{graphicx}
\usepackage{CJKutf8}
\usepackage{subfigure}

\usepackage{microtype}

\aclfinalcopy % Uncomment this line for the final submission
\def\aclpaperid{3136} %  Enter the acl Paper ID here

\title{Multimodal Joint Attribute Prediction and Value Extraction for E-commerce Product}

\author{Tiangang Zhu, Yue Wang, Haoran Li\thanks{\; Corresponding author.} , Youzheng Wu, Xiaodong He and Bowen Zhou \\ 
JD AI Research \\	
\texttt{\{zhutiangang3,wangyue274,lihaoran24,wuyouzheng1\}@jd.com}\\
\texttt{\{xiaodong.he,bowen.zhou\}@jd.com}
}

\date{}

\begin{document}\begin{CJK*}{UTF8}{gbsn}
\maketitle

\begin{abstract}
Product attribute values are essential in many e-commerce scenarios, such as customer service robots, product recommendations, and product retrieval. While in the real world, the attribute values of a product are usually incomplete and vary over time, which greatly hinders the practical applications. In this paper, we propose a multimodal method to jointly predict product attributes and extract values from textual product  descriptions with the help of the product images.
We argue that product attributes and values are highly correlated, \emph{e.g.}, it will be easier to extract the values on condition that the product attributes are given. Thus, we jointly model the attribute prediction and value extraction tasks from multiple aspects towards the interactions between attributes and values.
Moreover, product images have distinct effects on our tasks for different product attributes and values. Thus, we selectively draw useful visual information from product images to enhance our model.
We annotate a multimodal product attribute value dataset that contains 87,194 instances, and the experimental results on this dataset demonstrate that
explicitly modeling the relationship between attributes and values facilitates our method to establish the correspondence between them, and selectively utilizing visual product  information is necessary for the task.
Our code and dataset are available\footnote{https://github.com/jd-aig/JAVE}.
\end{abstract}

\begin{figure}
\centering
\includegraphics[width=0.95\linewidth]{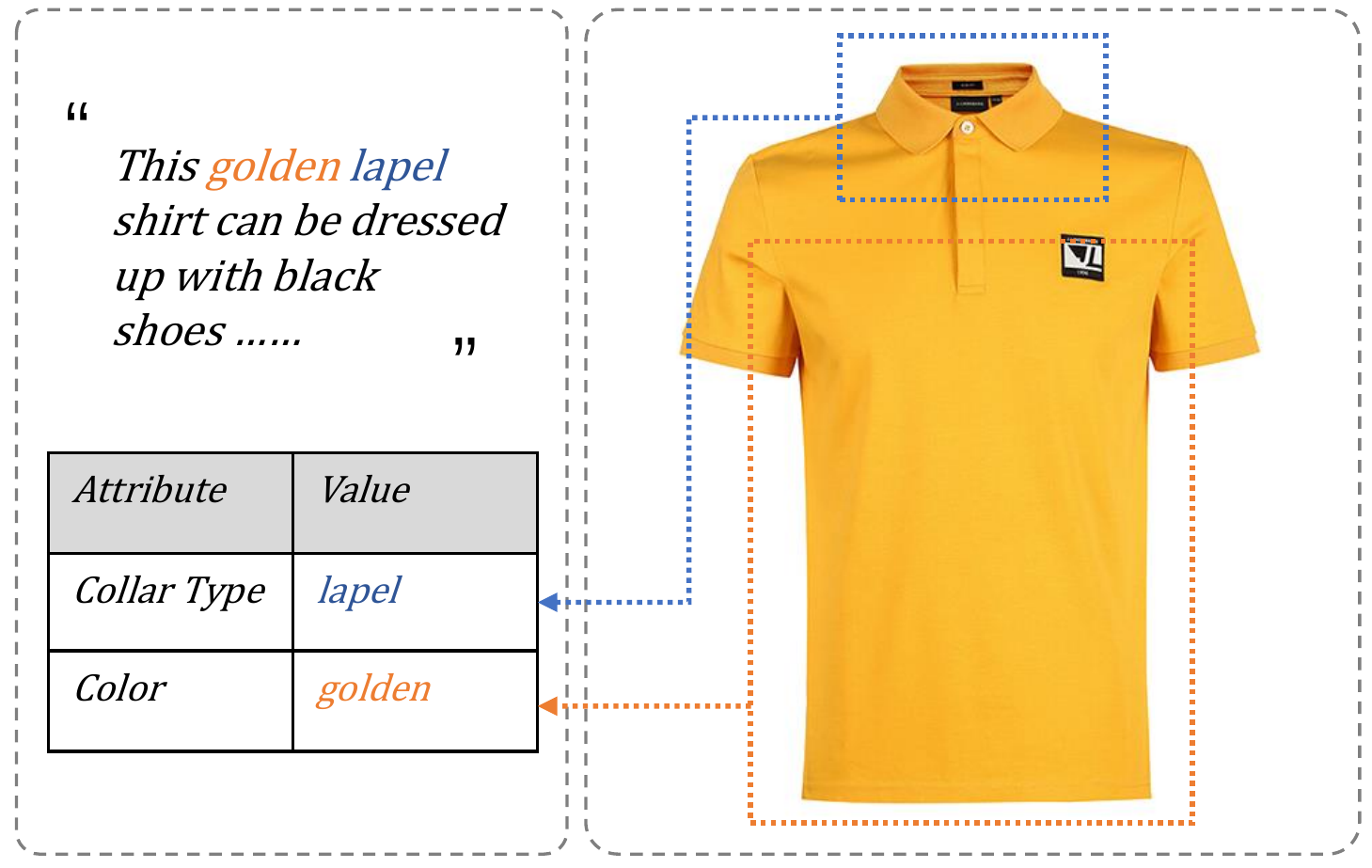}
\caption{An example of predicting attributes and extracting values from the textual product description with the aid of the visual product information.}
\label{pic:an_example}
\end{figure}

\section{Introduction}

Product attribute values that provide details of the product are crucial parts of e-commerce, which help customers to make purchasing decisions and facilitate retailers on many applications, such as question answering system~\cite{yih-etal-2015-semantic,yu-etal-2017-improved}, product recommendations~\cite{Gong09,CaoZGL18}, and product retrieval~\cite{Liao0ZNC18,MagnaniLXB19}.
While product attribute values are pervasively incomplete for a massive number of products on the e-commerce platform.
According to our statistics on a mainstream e-commerce platform in China, there are over 40 attributes for the products in \emph{clothing} category, but the average count of attributes present for each product is fewer than 8. The absence of the product attributes seriously  affects customers' shopping experience  and reduces  the potential of successful trading.
In this paper, we propose a method to jointly predict product attributes and extract the corresponding values with multimodal product information, as shown in Figure~\ref{pic:an_example}.

Though plenty of systems have been proposed to supplement product attribute values~\cite{PutthividhyaH11,More16,ShinzatoS13, ZhengMD018,XuWMJL19}, the relationship between product attributes and values are not sufficiently explored, and most of these approaches primarily focus on the text information. 
Attributes and values are, however, known to strongly
depend on each other, and vision can play a particularly essential role for this task.

\begin{figure*}
\centering
\includegraphics[width=1 \linewidth]{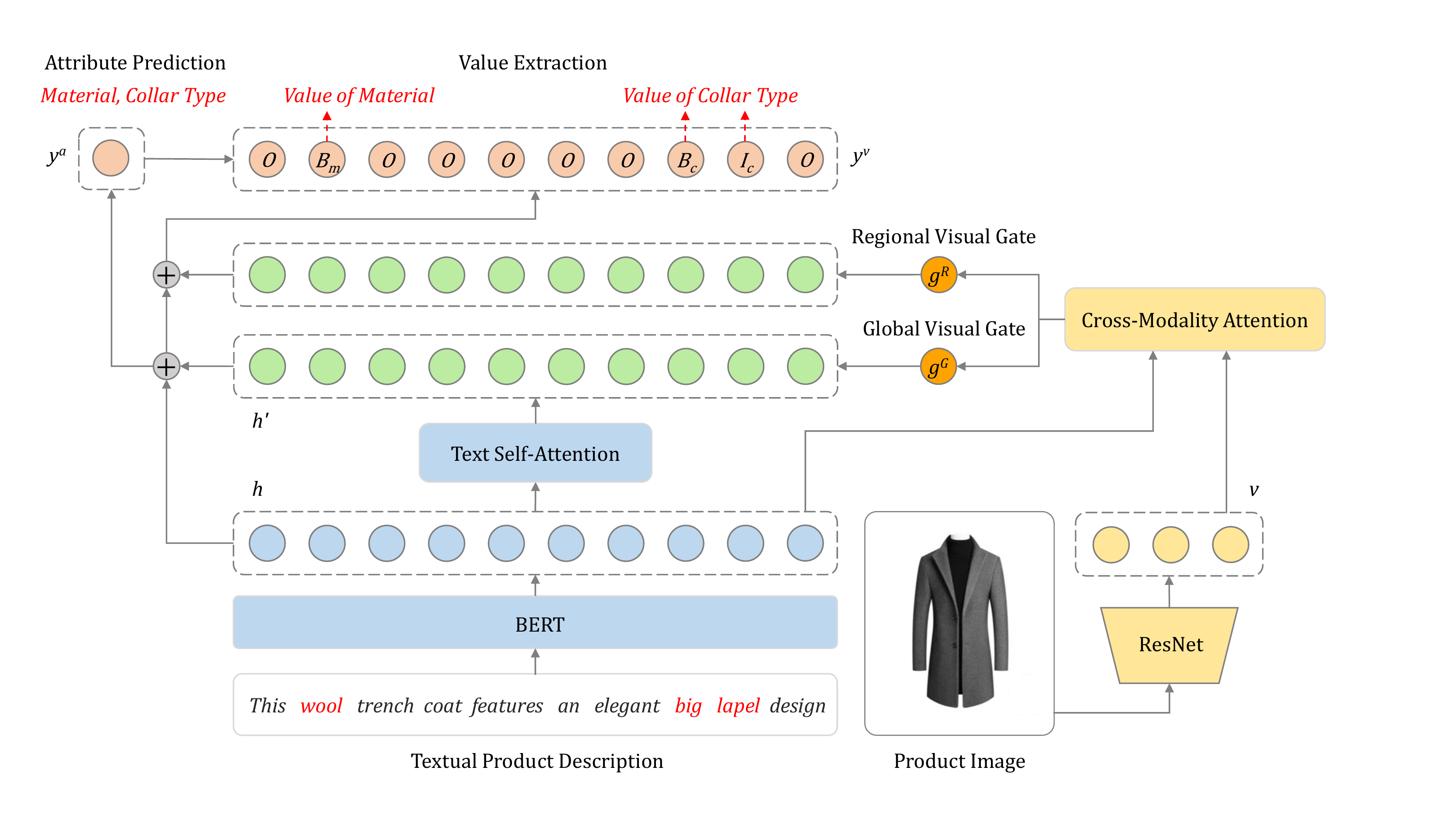}
\caption{Framework of our model.}
\label{pic:model_architecture}
\end{figure*}

Intuitively, product attributes and values are mutually indicative. Given a textual product description, we can extract attribute values more accurately with a known product attribute. We model the relationship between product attributes and values from the following three aspects.
First, we apply a multitask learning~\cite{Caruana97} method to  predict the product attributes and the values jointly. Second, we extract values with the guidance of the predicted product attributes. Third, we adopt a Kullback-Leibler (KL)~\cite{Kullback51klDivergence} measurement to penalize the inconsistency between the distribution of the product attribute prediction and that of the value extraction.

Furthermore, beyond the textual product descriptions, product images can provide additional clues for the attribute prediction and value extraction tasks. 
Figure~\ref{pic:an_example} illustrates this phenomenon. Given a description ``\emph{This golden band collar shirt can be dressed up with black shoes}", the term ``\emph{golden}" can be ambiguous for predicting the product attributes. While by viewing the product image, we can easily recognize the attribute corresponding to ``\emph{golden}" is ``\emph{Color}" instead of ``\emph{Material}". Moreover, the product image can indicate that the term ``\emph{black}" is not an attribute value of the current product; thus, it should not be extracted. This may be tricky for the model based on purely textual descriptions, but leveraging the visual information can make it easier. 
In addition, multimodal information shows promising efficiency on many tasks~\cite{LuYBP16,li-etal-2017-multi,BT0GZ18,LiZLZZ18,Yu0CT019,LiZMZZ19,tan-bansal-2019-lxmert,liu-etal-2019-graph,SuZCLLWD20,LiYXWHZ20}.
Therefore, we propose to incorporate visual information into our task. 
First, we selectively enhance the semantic representation of the textual product descriptions with a global-gated cross-modality attention module that is anticipated to benefit attribute prediction task with visually grounded semantics. 
Moreover, for different values, our model selectively utilizes visual information with a regional-gated cross-modality attention module to improve the accuracy of values extraction.

Our main contributions are threefold:
\begin{itemize}
\item We propose an end-to-end model to predict product attributes and extract the corresponding values.
\item Our model can selectively adopt visual product information by global and regional visual gates to enhance the attribute prediction and value extraction model.
\item We build a multimodal product attribute value dataset that contains 87,194 instances, involving various product categories.
\end{itemize}

\section{Model}
\subsection{Overview}

In this work, we tackle the product attribute-value pair completion task, \emph{i.e.}, predicting attributes and extracting the corresponding values for e-commerce products. The input of the task is a ``\emph{textual product description, product image}" pair, and the outputs are the product attributes (there may be more than one attribute in the descriptions) and the corresponding values. 
We model the product attribute prediction task as a sequence-level multilabel classification task and the value extraction task as a sequence labeling task. 

The framework of our proposed \underline{M}ultimodal \underline{J}oint \underline{A}ttribute Prediction and \underline{V}alue \underline{E}xtraction model (M-JAVE) is shown in Figure~\ref{pic:model_architecture}. The input sentence is encoded by a pretrained BERT model~\cite{DevlinCLT19}, and the image is encoded by a pretrained ResNet model~\cite{HeZRS16}. 
The global-gated cross-modality attention layer encodes text and image into the multimodal hidden representations.
Then, the M-JAVE model predicts the product attributes based on the multimodal representations. Next, the model extracts the values based on the previously predicted product attributes and the multimodal representations obtained through the regional-gated cross-modality attention layer. 
We apply the multitask learning framework to jointly model  the product attribute prediction and value extraction.
Considering the constraints between the product attributes and values, we adopt a KL loss to penalize the inconsistency between the distribution of the product attribute prediction and that of the value extraction.

\subsection{Text Encoder}
The text embedding vectors are encoded by a BERT-base model, which uses a concatenation of WordPiece~\cite{WuSCLNMKCGMKSJL16} embeddings, positional embeddings, and segment embeddings as the input representation. 
In addition, a special classification embedding (${[CLS]}$) is inserted as the first token, and a special token (${[SEP]}$) is added as the final token. Given a textual product description sentence decorated with two special tokens ${\textbf{x} = ([CLS], x_1, . . . , x_N, [SEP])}$, BERT outputs an embedding sequence $\textbf{h} = (h_0, h_1, . . . , h_N, h_{N+1})$. 

\subsection{Image Encoder}
We apply the ResNet~\cite{HeZRS16} to encode the product images.
We extract the activations from the last pooling layer of ResNet-101 that is pretrained  on the ImageNet~\cite{DengDSLL009} as the global visual feature $v_G$.
We use the $7\times7\times2048$ feature map of the $conv_5$ layer as the regional image feature $\textbf{v}=(v_1,...,v_K)$, where $K=49$.

\subsection{Global-Gated Cross-Modality Attention Layer}
Intuitively, for a specific product, as different modalities are semantically pertinent, we apply a cross-modality attention module to incorporate the textual and visual semantics into the multimodal hidden representations. 

Inspired by the self-attention mechanism~\cite{VaswaniSPUJGKP17}, we build a cross-modality attention layer capable of directly associating source tokens at different positions of the sentence and different regions of the image, by computing the attention score between each token-token pair and token-region pair, respectively. 
We argue that what is crucial to the cross-modality attention layer is the ability to selectively enrich the semantic representation of a sentence through the aid of an image. In other words, we need to avoid introducing noises resulted from when the image fails to represent some semantic meaning of words, such as abstract concepts. To achieve this, we design a global visual gate to filter out visual noise for any words that are irrelevant based on the visual signals.

Specifically, we feed the text and image representations ${h_{i}}$ and ${v_{k}}$ into the global-gated cross-modality attention layer, and then we obtain the  enhanced multimodal representation $ {h^{'}_{i}}$ as follows:
\begin{align}
e^{t}_{ij}&=(W_{Q}^{t}h_{i})(W_{K}^{t} h_{j})^{T}/{\sqrt{d}}\\
\alpha^{t}_{ij}&={\rm exp}(e^{t}_{ij})/\sum\nolimits_{m}{\rm exp}(e^{t}_{im})\\
e^{v}_{ik}&=(W_{Q}^{v}h_{i})(W_{K}^{v} v_{k})^{T}/{\sqrt{d}}\\
\alpha^{v}_{ik}&={\rm exp}(e^{v}_{ik})/\sum\nolimits_{n}{\rm exp}(e^{v}_{in})\\
{h^{'}_{i}} =& \sum\nolimits_{j}\alpha^{t}_{ij} W_{V}^{t} h_{j} +g_i^{G} \sum\nolimits_{k}\alpha^{v}_{ik}W_{V}^{v} v_{k}\label{eq:multimodal_resp}
\end{align}
where $W_{Q}^{t}$, $W_{K}^{t}$, $W_{V}^{t}$, $W_{Q}^{v}$, $W_{K}^{v}$, $W_{V}^{v}$ are weight matrices, and
$d$ is the dimension of $W_{Q}^{t}h_{i}$. 

The global visual gate $g_i^{G}$ is determined by the representation of the sentence and the image, which are obtained by the text encoder and the image encoder, respectively, as follows:
\begin{align}
g_i^G = \sigma (W_1h_i + W_2v_G + b)
\end{align}
where $W_1$ and $W_2$ are weight matrices.

\subsection{Product Attribute Prediction}
For an instance in the dataset, given ${\textbf{y}^a}=({y_1^a},...,{y_L^a})$, where ${{y_l^a}=1}$ denotes the instance with ${l}$-th attribute label, to predict the product attributes, we feed the text representation ${h_{i}}$, the multimodal representation ${h^{'}_{i}}$, and ${h_{0}}$ perceptron (the special classification element
${[CLS]}$ in BERT) into a feed-forward layer to output the predicted attribute labels $\hat{\textbf{y}^a}=(\hat{y_1}^a,...,\hat{y_L}^a)$:
\begin{align}
 \hat{\textbf{y}^a} = \sigma({W_{3}} \sum\nolimits_{i} {h_{i}} + {W_{4}} \sum\nolimits_{i} {h_{i}^{'}} + {W_{5}}{h_{0}})
\end{align}
where $W_3$, $W_4$ and $W_5$ are weight matrices.

Then we calculate the loss of the attribute prediction task by binary cross entropy over all $L$ labels:
\begin{align}
    {\rm Loss}_{a} = {\rm CrossEntropy}({\textbf{y}^a}, \hat{\textbf{y}^a})
\end{align}

\subsection{Product Value Extraction}
We regard the value extraction as a sequence labeling task that tags the input $\textbf{x}=(x_1,...,x_N)$ with the label sequence ${\textbf{y}^v}=(y_1^v,...,y_N^v)$ in the ${BIO}$
format,  \emph{e.g.}, attribute label ``\emph{Material}" corresponds to tags ``\emph{B-Material}" and ``\emph{I-Material}".

We argue that the product attributes can provide crucial indications for the attribute values. For example, given a sentence ``\emph{The red collar and golden buttons in the shirt form a colorful fashion topic}" and the predicted product attribute ``\emph{Color}", it is easy to recognize the value ``\emph{golden}" corresponding to attribute ``\emph{Color}" instead of ``\emph{Material}".
Thus, we incorporate the result of the product attribute prediction $\hat{\textbf{y}^a}$ to improve the value extraction.

Moreover, for a given product attribute, some regions of the image corresponding are more important than others. Thus, we set a gate $g^{R}_{k}$ for each image region to obtain a weighted visual semantic representation, which aims to use the regional image information more efficiently. 
Specifically, we feed text representation ${h_{i}}$, multimodal representation ${h^{'}_{i}}$, and image representation $v_k$ into a $\hbox{\textbf{regional-gated cross-modality attention layer}}$ and output the value labels $\hat{\textbf{y}^v}=(\hat{y_1}^v,...,\hat{y_N}^v)$:
\begin{align}
    \hat{{y}_{i}^{v}} =  {\rm softmax}(W_{6} h_{i} &+ W_{7}{h^{'}_{i}} + W_{8} \hat{\textbf{y}^a} \notag\\
    &+\sum\nolimits_{k}g^{R}_{k}\alpha^{v}_{ik}W_{V}^{v} v_{k})
\label{eq:pred_value}
\end{align}
where $W_6$, $W_7$, $W_8$, and $W_{V}^{v}$ are weight matrices.

The regional visual gate $g^{R}_{k}$ is determined by the regional visual semantics and the product attributes as follows:
\begin{align}
   g^{R}_{k} = \sigma(W_{9} \hat{\textbf{y}^a} + W_{10} v_{k})
\end{align}
where $W_9$ and $W_{10}$ are weight matrices.

Then we calculate the loss of the value extraction task by cross entropy:
\begin{align}
    {\rm Loss}_{v} = {\rm CrossEntropy}({\textbf{y}^v}, \hat{\textbf{y}^v})
\end{align}

\subsection{Multitask Learning}

To jointly model product attribute prediction and value extraction, our method is trained end-to-end via minimizing ${\rm Loss}_a$ and ${\rm Loss}_v$ coordinatively.

Moreover, the outputs of attribute prediction and value extraction are highly correlated, and thus we adopt a KL constraint between the outputs. Given the $l$-th attribute label, we assume that there are two corresponding value extraction tags \emph{e.g., attribute label ``Material" corresponds to tags ``B-Material" and ``I-Material"}, and their probabilities can be expressed as $\textbf{y}^v(B_{l})$ and $\textbf{y}^v(I_{l})$. Then the attribute prediction distribution mapped from the output of the corresponding value extraction task can be assigned as 
$\hat{\textbf{y}}^{v \rightarrow a}=(\hat{y}_1^{v \rightarrow a},...,\hat{y}_L^{v \rightarrow a})$, where 
\begin{align}
    {\hat{y}_l^{v \rightarrow a}} = \frac{1}{2}(\max\limits_{i}{\hat{y}_i^v(B_{l})} + \max\limits_{i}{\hat{y}_i^v(I_{l})})
\end{align}

The KL loss is:
\begin{align}
   {\rm KL}(\hat{\textbf{y}}^a||\hat{\textbf{y}}^{v \rightarrow a}) = \sum\nolimits_{l}{{\hat{{y}}_l^a}log\frac{\hat{{y}}_l^a}{\hat{{y}}_l^{v \rightarrow a}}}
   \label{KLloss}
\end{align}
and the final joint loss function is
\begin{align}
    {\rm Loss} = &  {\rm Loss}_{a} + {\rm Loss}_{v} 
     + \lambda {\rm KL}(\hat{\textbf{y}}^a||\hat{\textbf{y}}^{v \rightarrow a})
    \label{eq:total_loss}
\end{align}

\section{Dataset}
\label{our_dataset}
We collect a Multimodal E-commerce Product Attribute Value Extraction (MEPAVE) dataset with textual product descriptions and product images. 
Specifically, we collect instances from a mainstream Chinese e-commerce platform\footnote{https://www.jd.com/}. 
Crowdsourcing annotators are well-experienced in the area of e-commerce. Given a sentence, they are required to annotate the position of values mentioned in the sentence and label the corresponding attributes. 
In addition, the annotators also need to check the validity of the product text-image from its main page in e-commerce websites, and the unqualified ones will be removed.
We randomly select 1,000 instances to be annotated three times to ensure annotation consistency; the consistency rate is 92.83$\%$. 
Finally, we obtained 87,194 text-image instances consisting of the following categories of products: \emph{Clothes}, \emph{Pants}, \emph{Dresses}, \emph{Shoes}, \emph{Boots}, \emph{Luggage}, and \emph{Bags}, and involving 26 types of product attributes such as ``\emph{Material}", ``\emph{Collar Type}", ``\emph{Color}", etc. The distribution of different product categories and attribute values is shown in Table~\ref{tab:data_set}.
We randomly split all the instances into a training set with 71,194 instances, a validation set with 8,000 instances, and a testing set with 8,000 instances.

\begin{table}
\centering
\resizebox{.99\columnwidth}{!}{
\begin{tabular}{lcccc}
\hline \textbf{Category} & 
\textbf{$\#$Product} & \textbf{$\#$Instance} &
\textbf{$\#$Attr}&
\textbf{$\#$Value}
\\ 
\hline
Clothes & 12,240 & 34,154 & 14 & 1,210 \\
Shoes & 9,022 & 20,525 & 10 & 1,036 \\
Bags & 3,376 & 8,307 & 8 & 631 \\
Luggage & 1,291 & 2,227 & 7 & 275 \\
Dresses & 4,567 & 12,283 & 13 & 714 \\
Boots & 713 & 2,090 & 11 & 322 \\
Pants & 2,832 & 7,608 & 13 & 595 \\
\hline
Total & 34,041 & 87,194 & 26 & 2,129 \\
\hline
\end{tabular}
}
\caption{\label{font-table1} Statistics of the our dataset.}
\label{tab:data_set}
\end{table}

\citet{LoganHS17} release the English Multimodal Attribute Extraction (MAE) dataset. %collected from several e-commerce sites. 
Each instance in the MAE dataset contains a textual product description, a collection of images, and attribute-value pairs, where the values are not constrained to present in the textual product description. 
To verify our model on the MAE dataset, we select the instances whose values are in the textual product description, and we label the values by exactly matching. We denote this subset of the MAE dataset as MAE-text and the rest as MAE-image (values can be only inferred by the images).

\section{Experiment}

We compare our proposed methods with the following baselines:
\textbf{WSM} is the method that uses attribute values in the training set to retrieve the attribute values in the testing set by word matching.
\textbf{Sep-BERT} is the pretrained BERT model with feed-forward layers to perform these two subtasks separately.
\textbf{RNN-LSTM}~\cite{Hakkani-TurTCCG16}, \textbf{Attn-BiRNN}~\cite{LiuL16}, \textbf{Slot-Gated}~\cite{GooGHHCHC18}, and \textbf{Joint-BERT}~\cite{ChenBERT} are the  models to address intent classification and slot filling tasks, which are similar to the attribute prediction and value extraction, and we adopt these models to our task. 
\textbf{RNN-LSTM} and \textbf{Attn-BiRNN} use a bidirectional LSTM and an attention-based model for joint learning, respectively.
\textbf{Slot-Gated} introduces a gate-based mechanism to learn the relationship between these two tasks. 
\textbf{Joint-BERT} finetunes the BERT model with joint learning.
\textbf{ScalingUp}~\cite{XuWMJL19} adopts BiLSTM, CRF, and attention mechanism for introducing hidden semantic interaction between attribute and text. 

We report the results of our text-only and multimodal models, \emph{i.e.}, JAVE and M-JAVE. In addition, to eliminate the influences of different text encoders, we also conduct experiments with BiLSTM as the text encoder. 
Details about hyper-parameters are shown in Table~\ref{tab:hyper_parameters}.

\begin{table}\small
\centering
\resizebox{0.95\columnwidth}{!}{
\begin{tabular}{p{4cm}p{3cm}}
\hline
Item & Value \\ 
\hline 
Text Hidden Size & 768 \\
Image Hidden Size & 2048  \\
Image Block Number & 49 (7*7)  \\
Attention Vector Size& 200  \\
Max Sequence Length & 46 \\
Learning Rate & 0.0001 \\
Activation Function & sigmoid \\
Lambda for KL Loss & 0.5 \\
Batch Size & 128  \\
Epoch Number & 50 \\
Model Size & 112M  \\
GPU & 1x NVIDIA Tesla P40 \\
Training Time & 50 minutes\\
\hline
\end{tabular}
}
\caption{Details about hyper-parameters.}
\label{tab:hyper_parameters}
\end{table}

\begin{table}
\centering
\resizebox{1\columnwidth}{!}{
\begin{tabular}{lcc}
\hline \textbf{Model} & \textbf{Attribute} & \textbf{Value} \\ 
\hline
WSM & 77.20 & 72.52 \\
Sep-BERT & 86.34 & 83.12 \\
RNN-LSTM~\cite{Hakkani-TurTCCG16} & 85.76 & 82.92\\
Attn-BiRNN~\cite{LiuL16} & 86.10 & 83.28 \\
Slot-Gated~\cite{GooGHHCHC18} & 86.70 & 83.35\\
Joint-BERT~\cite{ChenBERT} & 86.93 & 83.73\\
ScalingUp~\cite{XuWMJL19} & - & 77.12 \\
\hline
JAVE (LSTM based) & 87.88 & 84.09\\
JAVE (BERT based) & 87.98 & 84.78\\
M-JAVE (LSTM based) & 90.19 & 86.41\\
M-JAVE (BERT based) & \textbf{90.69} & \textbf{87.17}\\
\hline
\end{tabular}
}
\caption{\label{font-table2} Main results ($\rm{{F}}_1$ score $\%$) of comparative methods and variants of our model.}
\label{tab:main_result}
\end{table}

\subsection{Main Results}

We evaluate our model on two subtasks, including attribute prediction and value extraction. The main results in Table~\ref{tab:main_result} show that our proposed M-JAVE model based on the BERT and the Bidirectional LSTM (BiLSTM)  both outperform the baselines significantly (paired t-test, p-value $< 0.01$), which proves an excellent generalization ability of our methods. 
From the results of our proposed M-JAVE and JAVE models, 
we can observe that the BERT is advantageous over the LSTM and visual product information improves the performance. 
The M-JAVE model achieves the best performance of 90.69$\%$ and 87.17$\%$ $F_{1}$ scores on two subtasks. 

Moreover, experimental results demonstrate the superiority of our JAVE model (either based on LSTM or BERT) against the models of \textbf{WSM}, \textbf{RNN-LSTM}, \textbf{Sep-BERT}, and joint-learning based models including \textbf{Attn-BiRNN}, \textbf{Slot-Gated} and \textbf{Joint-BERT}, indicating that the strategies for integrating the relationship between attributes and values into our models are necessary for the tasks.
We evaluate the \textbf{ScalingUp} model to predict the value for each given attribute on our dataset, and the result is unsatisfactory.
With the in-depth study, we found that 
it can be ascribed to identifying values that do not correspond to the given attribute. 
Over 34.52$\%$ of the predicted values are not the actual values for the input attributes, whereas this number is only 16.51$\%$ for our JAVE model. As a result, the \textbf{ScalingUp} model obtains a higher recall score (93.78$\%$) while a lower precision score of  (65.48$\%$) than our model (89.82$\%$ for recall score and 80.27$\%$ for the precision score). We argue that explicitly modeling the relationship between attributes and values facilitates our methods to establish the correspondence between them.

\begin{figure}
\centering
\includegraphics[width=0.95\linewidth]{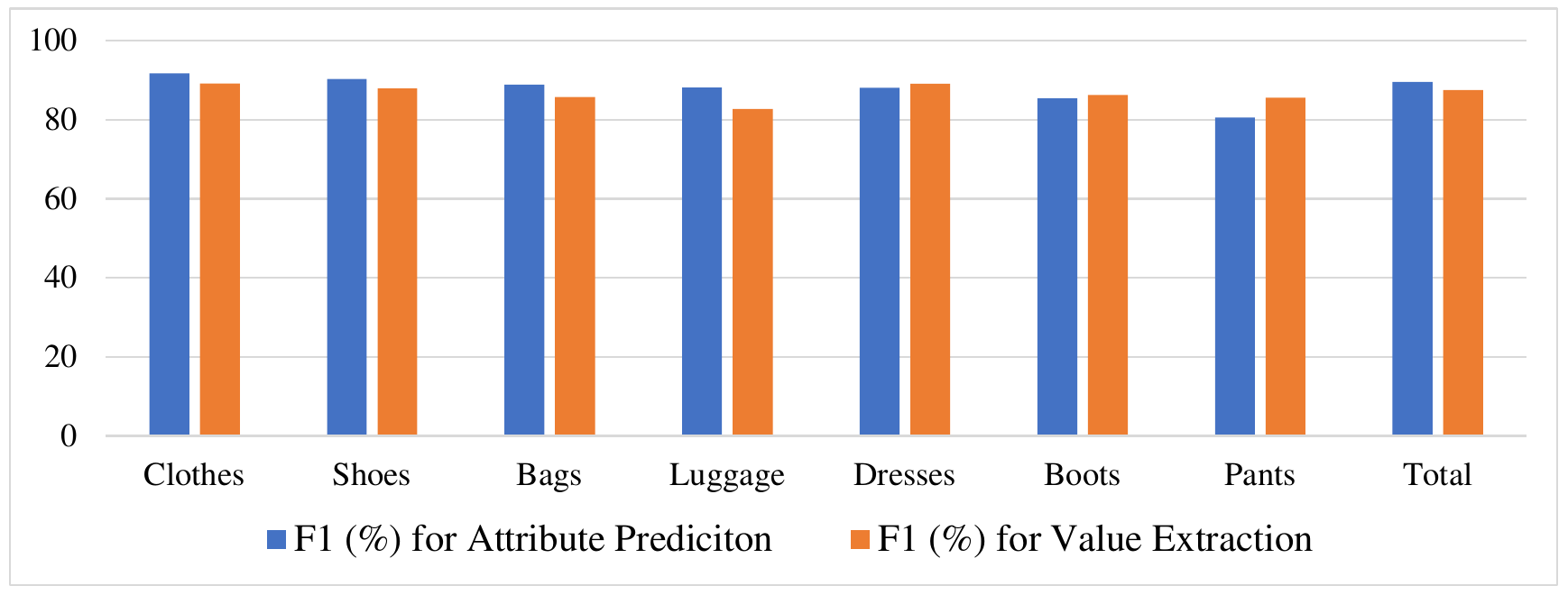}
\caption{Experimental results of the M-JAVE model for each product category.}
\label{pic:t1}
\end{figure}

\begin{figure*}
\centering
\includegraphics[width=0.99\linewidth]{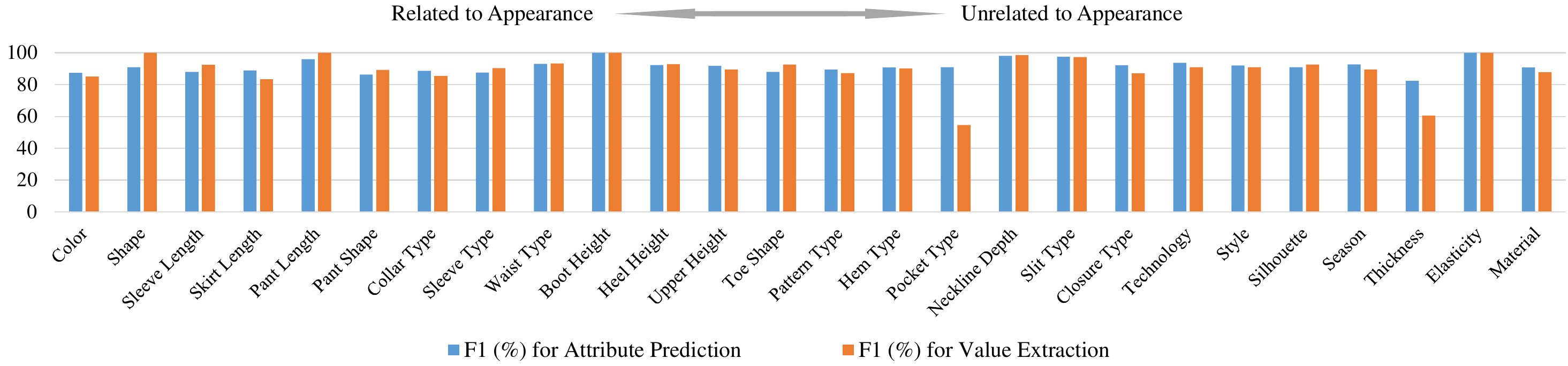}
\caption{Experimental results of the M-JAVE model for each type of attribute.}
\label{pic:t2}
\end{figure*}

More details including the results for each product category and for each type of attribute are shown in Figure~\ref{pic:t1} and \ref{pic:t2}.
We can find that our proposed method achieves satisfactory results for every category, and is not only suitable for simple attributes related to appearance, such as ``\emph{Color}" and ``\emph{Pant Length}", but also can deal with complex attributes, such as ``\emph{Elasticity}" and ``\emph{Material}".

To verify the adaptability of our proposed models, we conduct experiments on the English MAE dataset~\cite{LoganHS17}. The model proposed along with the MAE dataset  takes textual product  descriptions, visual information, and product attributes as input and treats the attribute value extraction task as predicting the value for a given product attribute. Thus, we compare our M-JAVE model with the MAE-model only on the value extraction task.

\begin{table}
\centering
\resizebox{.95\columnwidth}{!}{
\begin{tabular}{lccc}
\hline \textbf{Model} & 
\textbf{MAE}  &
\textbf{MAE-text}  &
\textbf{MAE-image}  \\
\hline
MAE-model & 59.48 & 72.96 & 52.11 \\
M-JAVE (LSTM) & - & 74.41 & - \\
M-JAVE (BERT) & - & 75.01 & - \\
\hline
\end{tabular}
}
\caption{\label{font-table3} Experimental results (accuracy $\%$) of our proposed model and MAE baseline model (MAE-model).}
\label{tab:compare_mae_result}
\end{table}

As shown in Table~\ref{tab:compare_mae_result}, on the MAE-text subset, our M-JAVE (LSTM) and M-JAVE (BERT) models outperform the MAE-model with 1.45$\%$ and 2.05$\%$ accuracy gains, respectively. On the original MAE and MAE-image subset, the accuracy scores of the MAE-model are 59.48$\%$ and 52.11$\%$, respectively, which are much lower than that on the MAE-text subset. We argue that it may be risky to predict the product values that do not appear in the textual product descriptions, and defining the value prediction as an extractive-based task is more reasonable for practical applications.

\subsection{Ablation Study} 
We perform ablation studies to confirm the effectiveness of the main modules of our models.

\begin{table}
\centering
\resizebox{.95\columnwidth}{!}{
\begin{tabular}{lcc}
\hline \textbf{Model} & \textbf{Attribute} & \textbf{Value} \\ \hline
\textbf{JAVE} & \textbf{87.98} & \textbf{84.78} \\
JAVE w/o MTL & 87.36 & 83.99 \\
JAVE w/o AttrPred & 86.74 & 83.90 \\
JAVE w/o KL-Loss & 87.24 & 84.26  \\
JAVE (UpBound of Attribute Task) & 89.03 & 100.0 \\
JAVE (UpBound of Value Task) & 100.0 & 88.72 \\
\hline
\end{tabular}
}
\caption{Experimental results ($F_{1}$ score $\%$) for ablation study on the relationship between attributes and values. ``UpBound" denotes ``Upper Bound".}
\label{tab:ablation_jave}
\end{table}

\subsubsection{Modeling the Relationship between Product Attributes and Values}

We explore the relationship between attributes and values from three aspects, including 1) applying the multitask learning to jointly predict the product attributes and values, 2) extracting values based on the predicted product attributes, and 3) introducing a KL loss to penalize the inconsistency between the results of product attributes and values.

Based on our text-only model, \emph{i.e.}, JAVE, we conduct experiments to evaluate the effectiveness of modeling the relationship by ablating the modules corresponding to the above three aspects.

\begin{itemize}
\item \emph{ w/o MTL} is the model without multitask learning (\emph{i.e.}, the two subtasks are addressed separately).
\item \emph{ w/o AttrPred} is the model without using the predicted product attributes in value extraction (\emph{i.e.}, remove ${W_8}\hat{\textbf{y}^a}$ in Eq.~\ref{eq:pred_value}).
\item \emph{ w/o KL loss} is the model without the KL loss (\emph{i.e.}, set $\lambda=0$ in Eq.~\ref{eq:total_loss}).
\end{itemize}

Furthermore, we get the upper bound of attribute prediction training with the ground-truth values (Eq.~\ref{KLloss}); we get the upper bound of value extraction training with the ground-truth attributes (Eq.~\ref{eq:pred_value} and \ref{KLloss}).

Table~\ref{tab:ablation_jave} shows a comparison of the JAVE model concerning the ablations. We can see that the JAVE model achieves the best performance. The results of the method ``JAVE w/o MTL", ``JAVE w/o AttrPred", and ``JAVE w/o KL loss" drop the $F_{1}$ scores by 0.62$\%$, 1.24$\%$, and 0.74$\%$ respectively for product attribute prediction, and drop the $F_{1}$ scores by 0.79$\%$, 0.88$\%$ and 0.52$\%$ respectively for value extraction, showing the effectiveness of modeling the relationship between product attributes and values. The results for the upper bound study shows the strong correlation between product attribute prediction and value extraction.

\begin{table}
\centering
\resizebox{1\columnwidth}{!}{
\begin{tabular}{lcc}
\hline \textbf{Model} & \textbf{Attribute} & \textbf{Value} \\ \hline
\textbf{M-JAVE} & \textbf{90.69} & \textbf{87.17}\\
M-JAVE w/o Visual Info (JAVE) & 87.98 & 84.78\\
M-JAVE w/o Global-Gated CrossMAtt  & 88.52 & 85.90 \\
M-JAVE w/o Regional-Gated CrossMAtt  & 88.29 & 85.38 \\
M-JAVE w/o Global Visual Gate & 87.27 & 80.32\\
M-JAVE w/o Regional Visual Gate & 87.66 & 82.54\\
\hline
\end{tabular}
}
\caption{\label{font-table5} 
Experimental results ($F_{1}$ score $\%$) for ablation study on the product images.}
\label{tab:ablation_mjave}
\end{table}

\subsubsection {Integrating Visual Product Information}

Our model mainly utilizes visual information of products from two aspects, including 1) predicting product attributes with a global-gated cross-modality attention module, and 2) extracting values with a regional-gated cross-modality attention module. We evaluate the effectiveness of visual product information as follows.

\begin{itemize}
\item \emph{ w/o Visual Info} is the model without utilizing visual information (\emph{i.e.}, JAVE).

\item \emph{ w/o Global-Gated CrossMAtt} is the model without the global-gated cross-modality attention (\emph{i.e.}, remove the right part in Eq.~\ref{eq:multimodal_resp}). 

\item \emph{ w/o Regional-Gated CrossMAtt} is the model without the regional-gated cross-modality attention (\emph{i.e.}, remove the right-most part in Eq.~\ref{eq:pred_value} inside the softmax function). 

\item \emph{ w/o Global Visual Gate} is the model without the global visual gate (\emph{i.e.}, remove $g_i^{G}$ in Eq.~\ref{eq:multimodal_resp}). 

\item \emph{ w/o Regional Visual Gate} is the model without the regional visual gate, (\emph{i.e.}, remove $g_k^{R}$ in Eq.~\ref{eq:pred_value}). 
\end{itemize}

From Table~\ref{tab:ablation_mjave}, we can see that removing global-gated or regional-gated cross-modality attention modules degrades the performances on both subtasks, proving the effectiveness of visual information for our task.

Moreover, for the models with cross-modality attention modules while without global or regional visual gates, \emph{i.e.}, M-JAVE w/o Global Visual Gate and M-JAVE w/o Regional Visual Gate, respectively, the performances are  worse than that of M-JAVE significantly. Remarkably, the models of M-JAVE w/o Global Visual Gate and M-JAVE w/o Regional Visual Gate underperform the models thoroughly removing visual-related modules.

To sum up, using the visual product  information indiscriminately poses detrimental  effects on the model, and selectively utilizing visual product information with global and regional visual gates are essential for our tasks. Further experiment about the visual information is in the Appendices.

\subsection{Adversarial Evaluation of Attribute Prediction and Value Extraction}
To further verify whether the visual product  information can improve the performance of product attribute prediction and value extraction, we adopt an adversarial evaluation method~\citep{Elliott18} that measures the performance variation when our model is presented with a random incongruent image.

The awareness score of a model $\mathcal{M}$ on an evaluation dataset $\mathcal{D}$ is defined as follows:
\begin{align}
    \Delta_{Awareness} = \frac{1}{\left | \mathcal{D} \right |}  \sum_{i}^{\left | \mathcal{D} \right |} a_{\mathcal{M}}(x_{i}, y_{i}, v_{i}, \bar{v}_{i})
\end{align}
Where $\Delta_{Awareness}$ denotes the image awareness. $\emph{x}$, $\emph{y}$ denote the the text  and the  values of the product, respectively. $\emph{v}$, $\bar{\emph{v}}$ denote the congruent image and the incongruent image, respectively.

We use the $F_{1}$ score to calculate awareness score for a single instance:
\begin{align}
    a_{\mathcal{M}} = F_{1}(x_{i}, y_{i}, v_{i}) - F_{1}(x_{i}, y_{i}, \bar{v}_{i}) \label{i_a score}
\end{align}
Under this definition, the output of the evaluation performance measure should be higher in the presence of the congruent data than incongruent data, i.e., ${F_{1}(x_{i}, y_{i}, v_{i}) > F_{1}(x_{i}, y_{i}, \bar{v}_{i})}$. If this is the case, on average, then the overall image awareness of a model ${\Delta_{Awareness}}$ is positive. This can only happen when model outputs are evaluated more favourably in the presence of the congruent image data than the incongruent image data.

To determine if a model passes the proposed evaluation, we conduct the statistical test using the pairs of values that are calculated in the process of computing the image awareness scores (Eq.~\ref{i_a score})

Table ~\ref{tab:awareness} shows the evaluation results of product attribute prediction and value extraction. We find that, on both subtasks, the $F_{1}$ scores with incongruent images are much lower than that with the congruent images, and ${\Delta_{Awareness}}$ is significant positive. Moreover, we use $K=8$ separate p values from each test based on Fisher's method, and get ${\mathcal{X}^2}$=6790.80, ${p<}$0.0001 in product attribute prediction and ${\mathcal{X}^2}$=780.80, $p<$0.0001 in value extraction, which proves that the incongruent image significantly degrades the model's performance. We can conclude that the visual information make substantial contribution to the attribute prediction and value extraction tasks.

\begin{table}
\centering
\resizebox{.99\columnwidth}{!}{
\begin{tabular}{lccc}
\hline  
& \textbf{C} & \textbf{I} & ${{\Delta}_{Awareness}}$ \\ 
\hline
\textbf{Value} & 87.48 & 78.57$_{0.23}$ & 11.26$_{0.18}$ \\
\textbf{Attribute} & 89.57 & 86.64$_{0.13}$ & 3.2$_{0.08}$ \\
\hline
\end{tabular}
}
\caption{\label{font-table4} $F_1$ scores in the \textbf{C}ongruent and \textbf{I}ncongruent settings, along with the Meteor-awareness results. Incongruent and ${\Delta_{Awareness}}$ scores are the mean and standard deviation of 8 permutations of product images in test dataset.}
\label{tab:awareness}
\end{table}

\subsection{Domain Adaptation}
To verify the robustness of our models, we conduct an evaluation on the out-of-domain data.
The source domain is our formal product information (PI) mentioned in Section~\ref{our_dataset}.
The target domain is the oral Question-Answering (QA), where the textual description consists of QA pairs about the product in the real e-commerce customer service dialogue,  and the visual information is from the image of product mentioned in the dialogue. We directly apply the JAVE and M-JAVE models trained on PI to test on the QA  testing set containing  900 manually annotated instances. 

As shown in Table~\ref{tab:domain_adaptation}, on the QA testing set, M-JAVE outperforms JAVE with 4.31$\%$ and 5.70$\%$ $F_{1}$ scores on the attribute prediction and value extraction tasks, respectively.
For the attribute prediction task, the gap between the results on the PI and QA testing reduces from 14.58$\%$ to 12.98$\%$ when using the visual information. 
Similarly, the gap reduces from 14.31$\%$ to 11.00$\%$ for the value extraction task.
This demonstrates that visual product  information makes our model more robust.

\begin{table}
\centering
\resizebox{.99\columnwidth}{!}{
\begin{tabular}{l|c|c|c|c|c|c}
\hline
 & \multicolumn{3}{|c}{\textbf{Attribute}} & \multicolumn{3}{|c}{\textbf{Value}} \\
\hline
\textbf{Models} & \textbf{PI} & \textbf{QA} & \textbf{${\Delta}_{\downarrow}$} & \textbf{PI} & \textbf{QA} & \textbf{${\Delta}_{\downarrow}$}\\ \hline
JAVE  & 87.98 & 73.40 & 14.58 & 84.78 & 70.47 & 14.31 \\
M-JAVE  & 90.69 & 77.71 & 12.98 & 87.17 & 76.17 & 11.00 \\
\hline
\end{tabular}
}
\caption{\label{font-table6} Experimental results ($F_{1}$ score $\%$) for domain adaptation. $\Delta_{\downarrow}$ denotes the $F_{1}$ score gap for the PI and QA domains.}
\label{tab:domain_adaptation}
\end{table}

\subsection{Low-Resource Evaluation}

\begin{table}
\centering
\resizebox{.99\columnwidth}{!}{
\begin{tabular}{l|c|c|c|c|c}
\hline
\textbf{\% of data} & \textbf{100\%} & \textbf{80\%} & \textbf{60\%} & \textbf{40\%} & \textbf{20\%} \\
\hline
\textbf{Attribute} \\
\hline
JAVE & 87.98 & 86.83$_{0.31}$ & 84.81$_{0.64}$ & 76.89$_{2.81}$ & 72.13$_{3.64}$\\ 
M-JAVE & 90.69 & 88.48$_{0.21}$ & 86.14$_{0.52}$ & 81.23$_{1.66}$ & 78.70$_{2.92}$\\
\hline
\textbf{Value} \\
\hline
JAVE & 84.78 & 82.77$_{0.45}$ & 78.81$_{0.82}$ & 74.12$_{2.42}$ & 66.57$_{4.24}$\\ 
M-JAVE & 87.17 & 86.61$_{0.28}$ & 83.88$_{0.67}$ & 79.67$_{1.87}$ & 74.63$_{3.23}$\\
\hline
\end{tabular}
}
\caption{\label{font-table} 
Results (mean and standard deviation) with different sizes of of training data.}
\label{tab:cutting_training_set}
\end{table}

To further verify the robustness of our model, we evaluate our models trained with subsets of the whole training set in different proportions. For each proportion, we randomly sample the training data three times, and we report the mean and standard deviation in Table~\ref{tab:cutting_training_set}.
It illustrates that visual product information brings 
considerable advantages on the robustness when few training instances are available. 

\begin{figure}\small
\centering
\includegraphics[width=0.85\linewidth]{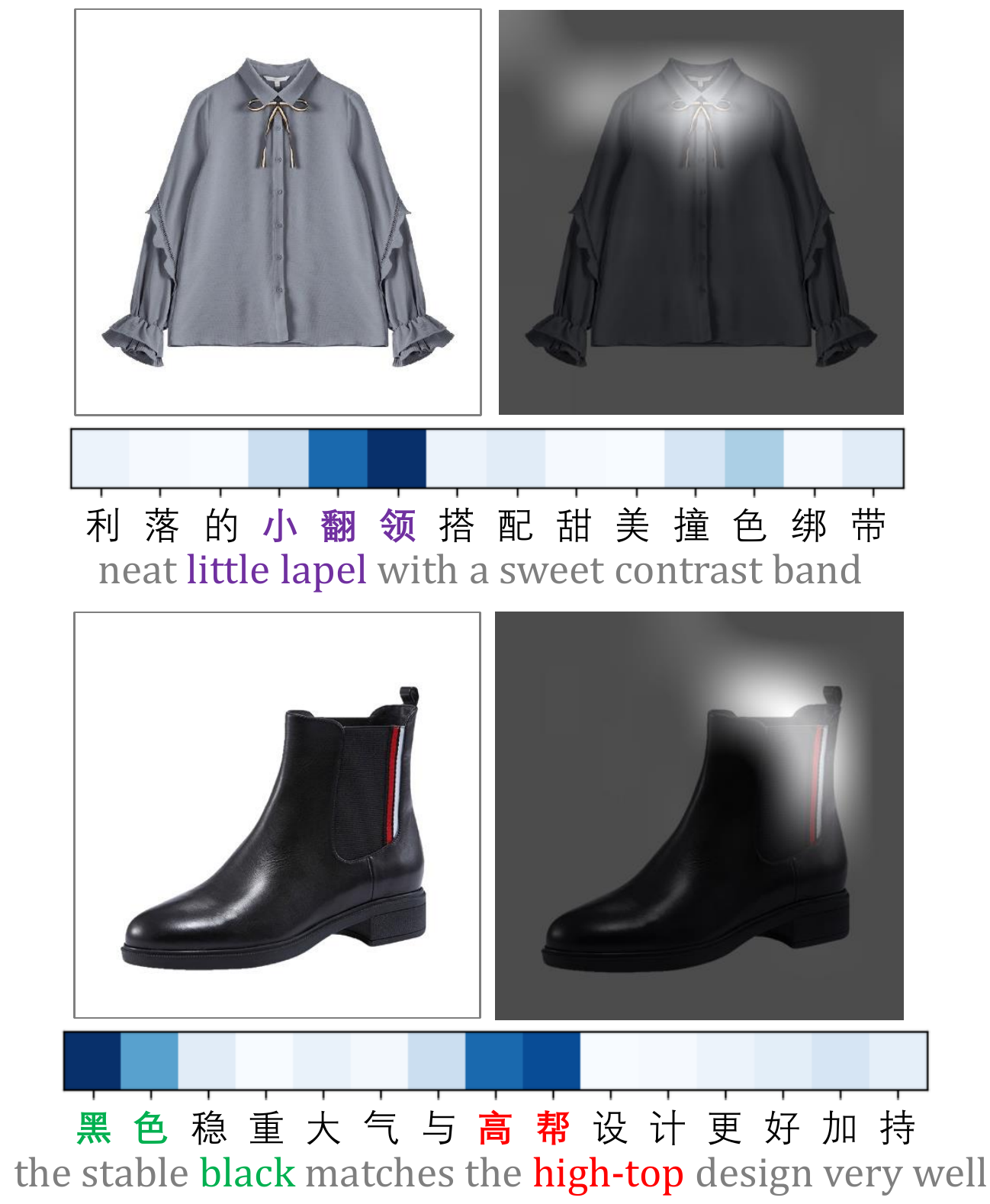}
\caption{Heat maps for global (blocks above the text) and regional (images on the right) visual gates.}
\label{pic:attn_image}
\end{figure}

\subsection{Visualization}
To evaluate the global and regional visual gates qualitatively, we visualize these gates for different attribute values with the M-JAVE model.
The results are shown in Figure~\ref{pic:attn_image}.
For the blocks above the text, the deeper color denotes the larger value for the global visual gate $g_i^G$, \emph{i.e.},  more visual information is used for enhancing the semantic meaning of the text.
We can find that the global visual gates are positively related to the relevance between the text and the image.
For the product image on the right, the lighter color denotes the larger value for the regional visual gate $g_k^R$, \emph{i.e.},  more visual information is drawn for extracting values.
The results demonstrate that the regional visual gate successfully captures useful parts of the product image.

\section{Related Work}
 
Recent approaches related to the attribute value pair completion task can be classified as the following two categories.

\textbf{1) Predicting  integral attribute-value tags.} 
\citet{PutthividhyaH11} and 
\citet{ZhengMD018}
introduce a set of entity tags for each attribute  (\emph{e.g.}, ``\emph{B-Material}" and ``\emph{I-Material}" for the attribute ``\emph{Material}"). \citet{PutthividhyaH11} adopt a NER system with bootstrapping to predict values, and \citet{ZhengMD018} apply a Bi-LSTM-CRF model with the attention mechanism. 
It may be challenging to handle the massive amounts of attributes in the real world.

\textbf{2) Predicting values for given attributes.} \citet{GhaniPLKF06}  treat the task as a value classification task and create a specific text classifier for each given attribute.  \citet{More16} and \citet{XuWMJL19}  formulate the task as a special case of NER~\cite{BikelSW99,CollobertWBKKK11} task that predicts the values for each attribute. 
\citet{More16} combines CRF and structured perceptron with a curated normalization scheme to predict values, and \citet{XuWMJL19} regard attributes as queries and adopt BIO tags for any attributes, making it applicable for the large-scaled attribute system.
However, our experimental results show that the model of \citet{XuWMJL19} may be insufficient to identify which attribute a value corresponds to.

In this paper, we propose a third category of method: \textbf{jointly predicting attributes and extracting values}. 
The attribute prediction module provides guidance and constraints for the value extraction module, which adapts our model to fit large-scaled attribute applications.
Moreover, we explicitly model the relationship between attributes and values,  which  helps to establish the correspondence between them effectively.

\section{Conclusion}
We jointly tackle the tasks of e-commerce product attribute prediction and value extraction from multiple aspects towards the relationship between product attributes and values, and we prove that the models can benefit a lot from visual product information.
The experimental results show that the correlations between product attributes and values are valuable for this task, and visual information should be selectively used.
%We construct a multimodal product attribute value dataset. The dataset and the code are available\footnote{https://github.com/wavewangyue/jave}.

\bibliography{emnlp2020}

\begin{thebibliography}{37}
\expandafter\ifx\csname natexlab\endcsname\relax\def\natexlab#1{#1}\fi

\bibitem[{Anderson et~al.(2018)Anderson, He, Buehler, Teney, Johnson, Gould,
  and Zhang}]{BT0GZ18}
Peter Anderson, Xiaodong He, Chris Buehler, Damien Teney, Mark Johnson, Stephen
  Gould, and Lei Zhang. 2018.
\newblock \href
  {http://openaccess.thecvf.com/content\_cvpr\_2018/html/Anderson\_Bottom-Up\_and\_Top-Down\_CVPR\_2018\_paper.html}
  {Bottom-up and top-down attention for image captioning and visual question
  answering}.
\newblock In \emph{2018 {IEEE} Conference on Computer Vision and Pattern
  Recognition, {CVPR} 2018, Salt Lake City, UT, USA, June 18-22, 2018}, pages
  6077--6086.

\bibitem[{Bikel et~al.(1999)Bikel, Schwartz, and Weischedel}]{BikelSW99}
Daniel~M. Bikel, Richard~M. Schwartz, and Ralph~M. Weischedel. 1999.
\newblock \href {https://doi.org/10.1023/A:1007558221122} {An algorithm that
  learns what's in a name}.
\newblock \emph{Mach. Learn.}, 34(1-3):211--231.

\bibitem[{Cao et~al.(2018)Cao, Zhou, Gao, and Li}]{CaoZGL18}
Min Cao, Sijing Zhou, Honghao Gao, and Youhuizi Li. 2018.
\newblock \href {https://doi.org/10.18293/SEKE2018-050} {A novel hybrid
  collaborative filtering approach to recommendation using reviews: The product
  attribute perspective {(S)}}.
\newblock In \emph{The 30th International Conference on Software Engineering
  and Knowledge Engineering, Hotel Pullman, Redwood City, California, USA, July
  1-3, 2018}, pages 7--10.

\bibitem[{Caruana(1997)}]{Caruana97}
Rich Caruana. 1997.
\newblock \href {https://link.springer.com/article/10.1023/A:1007379606734}
  {Multitask learning}.
\newblock \emph{Machine learning}, 28(1):41--75.

\bibitem[{Chen et~al.(2019)Chen, Zhuo, and Wang}]{ChenBERT}
Qian Chen, Zhu Zhuo, and Wen Wang. 2019.
\newblock \href {http://arxiv.org/abs/1902.10909} {{BERT} for joint intent
  classification and slot filling}.
\newblock \emph{CoRR}, abs/1902.10909.

\bibitem[{Collobert et~al.(2011)Collobert, Weston, Bottou, Karlen, Kavukcuoglu,
  and Kuksa}]{CollobertWBKKK11}
Ronan Collobert, Jason Weston, L{\'{e}}on Bottou, Michael Karlen, Koray
  Kavukcuoglu, and Pavel~P. Kuksa. 2011.
\newblock \href {http://dl.acm.org/citation.cfm?id=2078186} {Natural language
  processing (almost) from scratch}.
\newblock \emph{J. Mach. Learn. Res.}, 12:2493--2537.

\bibitem[{Deng et~al.(2009)Deng, Dong, Socher, Li, Li, and Li}]{DengDSLL009}
Jia Deng, Wei Dong, Richard Socher, Li{-}Jia Li, Kai Li, and Fei{-}Fei Li.
  2009.
\newblock \href {https://doi.org/10.1109/CVPR.2009.5206848} {Image{N}et: {A}
  large-scale hierarchical image database}.
\newblock In \emph{2009 {IEEE} Computer Society Conference on Computer Vision
  and Pattern Recognition {(CVPR} 2009), 20-25 June 2009, Miami, Florida,
  {USA}}, pages 248--255.

\bibitem[{Devlin et~al.(2019)Devlin, Chang, Lee, and Toutanova}]{DevlinCLT19}
Jacob Devlin, Ming-Wei Chang, Kenton Lee, and Kristina Toutanova. 2019.
\newblock \href {https://doi.org/10.18653/v1/N19-1423} {{BERT}: Pre-training of
  deep bidirectional transformers for language understanding}.
\newblock In \emph{Proceedings of the 2019 Conference of the North {A}merican
  Chapter of the Association for Computational Linguistics}, pages 4171--4186,
  Minneapolis, Minnesota.

\bibitem[{Elliott(2018)}]{Elliott18}
Desmond Elliott. 2018.
\newblock \href {https://www.aclweb.org/anthology/D18-1329} {Adversarial
  evaluation of multimodal machine translation}.
\newblock In \emph{Proceedings of the 2018 Conference on Empirical Methods in
  Natural Language Processing, Brussels, Belgium, October 31 - November 4,
  2018}, pages 2974--2978.

\bibitem[{Ghani et~al.(2006)Ghani, Probst, Liu, Krema, and Fano}]{GhaniPLKF06}
Rayid Ghani, Katharina Probst, Yan Liu, Marko Krema, and Andrew~E. Fano. 2006.
\newblock \href {https://doi.org/10.1145/1147234.1147241} {Text mining for
  product attribute extraction}.
\newblock \emph{{SIGKDD} Explorations}, 8(1):41--48.

\bibitem[{Gong(2009)}]{Gong09}
SongJie Gong. 2009.
\newblock \href {https://doi.org/10.4304/jsw.4.8.883-890} {Employing user
  attribute and item attribute to enhance the collaborative filtering
  recommendation}.
\newblock \emph{{JSW}}, 4(8):883--890.

\bibitem[{Goo et~al.(2018)Goo, Gao, Hsu, Huo, Chen, Hsu, and
  Chen}]{GooGHHCHC18}
Chih-Wen Goo, Guang Gao, Yun-Kai Hsu, Chih-Li Huo, Tsung-Chieh Chen, Keng-Wei
  Hsu, and Yun-Nung Chen. 2018.
\newblock \href {https://doi.org/10.18653/v1/N18-2118} {Slot-gated modeling for
  joint slot filling and intent prediction}.
\newblock In \emph{Proceedings of the 2018 Conference of the North {A}merican
  Chapter of the Association for Computational Linguistics: Human Language
  Technologies, Volume 2 (Short Papers)}, pages 753--757, New Orleans,
  Louisiana.

\bibitem[{Hakkani{-}T{\"{u}}r et~al.(2016)Hakkani{-}T{\"{u}}r, T{\"{u}}r,
  {\c{C}}elikyilmaz, Chen, Gao, Deng, and Wang}]{Hakkani-TurTCCG16}
Dilek Hakkani{-}T{\"{u}}r, G{\"{o}}khan T{\"{u}}r, Asli {\c{C}}elikyilmaz,
  Yun{-}Nung Chen, Jianfeng Gao, Li~Deng, and Ye{-}Yi Wang. 2016.
\newblock \href {https://doi.org/10.21437/Interspeech.2016-402} {Multi-domain
  joint semantic frame parsing using bi-directional {RNN-LSTM}}.
\newblock In \emph{Interspeech 2016, 17th Annual Conference of the
  International Speech Communication Association, San Francisco, CA, USA,
  September 8-12, 2016}, pages 715--719.

\bibitem[{He et~al.(2016)He, Zhang, Ren, and Sun}]{HeZRS16}
Kaiming He, Xiangyu Zhang, Shaoqing Ren, and Jian Sun. 2016.
\newblock \href {https://doi.org/10.1109/CVPR.2016.90} {Deep residual learning
  for image recognition}.
\newblock In \emph{2016 {IEEE} Conference on Computer Vision and Pattern
  Recognition, {CVPR} 2016, Las Vegas, NV, USA, June 27-30, 2016}, pages
  770--778.

\bibitem[{IV et~al.(2017)IV, Humeau, and Singh}]{LoganHS17}
Robert L.~Logan IV, Samuel Humeau, and Sameer Singh. 2017.
\newblock \href {http://arxiv.org/pdf/1711.11118} {Multimodal attribute
  extraction}.
\newblock In \emph{6th Workshop on Automated Knowledge Base Construction,
  AKBC@NIPS 2017, Long Beach, California, USA, December 8, 2017}.

\bibitem[{Kullback and Leibler(1951)}]{Kullback51klDivergence}
Solomon Kullback and Richard~A Leibler. 1951.
\newblock \href {https://www.jstor.org/stable/2236703} {On information and
  sufficiency}.
\newblock \emph{The annals of mathematical statistics}, 22(1):79--86.

\bibitem[{Li et~al.(2020)Li, Yuan, Xu, Wu, He, and Zhou}]{LiYXWHZ20}
Haoran Li, Peng Yuan, Song Xu, Youzheng Wu, Xiaodong He, and Bowen Zhou. 2020.
\newblock \href {https://aaai.org/ojs/index.php/AAAI/article/view/6332}
  {Aspect-aware multimodal summarization for chinese e-commerce products}.
\newblock In \emph{The Thirty-Fourth {AAAI} Conference on Artificial
  Intelligence, {AAAI} 2020}, pages 8188--8195.

\bibitem[{Li et~al.(2018)Li, Zhu, Liu, Zhang, and Zong}]{LiZLZZ18}
Haoran Li, Junnan Zhu, Tianshang Liu, Jiajun Zhang, and Chengqing Zong. 2018.
\newblock \href {https://doi.org/10.24963/ijcai.2018/577} {Multi-modal sentence
  summarization with modality attention and image filtering}.
\newblock In \emph{Proceedings of the Twenty-Seventh International Joint
  Conference on Artificial Intelligence, {IJCAI} 2018, July 13-19, 2018,
  Stockholm, Sweden}, pages 4152--4158.

\bibitem[{Li et~al.(2017)Li, Zhu, Ma, Zhang, and Zong}]{li-etal-2017-multi}
Haoran Li, Junnan Zhu, Cong Ma, Jiajun Zhang, and Chengqing Zong. 2017.
\newblock \href {https://doi.org/10.18653/v1/D17-1114} {Multi-modal
  summarization for asynchronous collection of text, image, audio and video}.
\newblock In \emph{Proceedings of the 2017 Conference on Empirical Methods in
  Natural Language Processing}, pages 1092--1102, Copenhagen, Denmark.

\bibitem[{Li et~al.(2019)Li, Zhu, Ma, Zhang, and Zong}]{LiZMZZ19}
Haoran Li, Junnan Zhu, Cong Ma, Jiajun Zhang, and Chengqing Zong. 2019.
\newblock \href {https://doi.org/10.1109/TKDE.2018.2848260} {Read, watch,
  listen, and summarize: Multi-modal summarization for asynchronous text,
  image, audio and video}.
\newblock \emph{{IEEE} Transactions on Knowledge and Data Engineering},
  31(5):996--1009.

\bibitem[{Liao et~al.(2018)Liao, He, Zhao, Ngo, and Chua}]{Liao0ZNC18}
Lizi Liao, Xiangnan He, Bo~Zhao, Chong{-}Wah Ngo, and Tat{-}Seng Chua. 2018.
\newblock \href {https://doi.org/10.1145/3240508.3240646} {Interpretable
  multimodal retrieval for fashion products}.
\newblock In \emph{2018 {ACM} Multimedia Conference on Multimedia Conference,
  {MM} 2018, Seoul, Republic of Korea, October 22-26, 2018}, pages 1571--1579.

\bibitem[{Liu and Lane(2016)}]{LiuL16}
Bing Liu and Ian Lane. 2016.
\newblock \href {https://doi.org/10.21437/Interspeech.2016-1352}
  {Attention-based recurrent neural network models for joint intent detection
  and slot filling}.
\newblock In \emph{Interspeech 2016, 17th Annual Conference of the
  International Speech Communication Association, San Francisco, CA, USA,
  September 8-12, 2016}, pages 685--689.

\bibitem[{Liu et~al.(2019)Liu, Gao, Zhang, and Zhao}]{liu-etal-2019-graph}
Xiaojing Liu, Feiyu Gao, Qiong Zhang, and Huasha Zhao. 2019.
\newblock \href {https://doi.org/10.18653/v1/N19-2005} {Graph convolution for
  multimodal information extraction from visually rich documents}.
\newblock In \emph{Proceedings of the 2019 Conference of the North {A}merican
  Chapter of the Association for Computational Linguistics: Human Language
  Technologies, Volume 2 (Industry Papers)}, pages 32--39, Minneapolis,
  Minnesota.

\bibitem[{Lu et~al.(2016)Lu, Yang, Batra, and Parikh}]{LuYBP16}
Jiasen Lu, Jianwei Yang, Dhruv Batra, and Devi Parikh. 2016.
\newblock \href
  {http://papers.nips.cc/paper/6202-hierarchical-question-image-co-attention-for-visual-question-answering}
  {Hierarchical question-image co-attention for visual question answering}.
\newblock In \emph{Advances in Neural Information Processing Systems 29: Annual
  Conference on Neural Information Processing Systems 2016, December 5-10,
  2016, Barcelona, Spain}, pages 289--297.

\bibitem[{Magnani et~al.(2019)Magnani, Liu, Xie, and Banerjee}]{MagnaniLXB19}
Alessandro Magnani, Feng Liu, Min Xie, and Somnath Banerjee. 2019.
\newblock \href {https://doi.org/10.1145/3308560.3316603} {Neural product
  retrieval at walmart.com}.
\newblock In \emph{Companion of The 2019 World Wide Web Conference, {WWW} 2019,
  San Francisco, CA, USA, May 13-17, 2019}, pages 367--372.

\bibitem[{More(2016)}]{More16}
Ajinkya More. 2016.
\newblock \href {http://arxiv.org/abs/1608.04670} {Attribute extraction from
  product titles in ecommerce}.
\newblock \emph{CoRR}, abs/1608.04670.

\bibitem[{Putthividhya and Hu(2011)}]{PutthividhyaH11}
Duangmanee Putthividhya and Junling Hu. 2011.
\newblock \href {https://www.aclweb.org/anthology/D11-1144} {Bootstrapped named
  entity recognition for product attribute extraction}.
\newblock In \emph{Proceedings of the 2011 Conference on Empirical Methods in
  Natural Language Processing}, pages 1557--1567, Edinburgh, Scotland, UK.

\bibitem[{Shinzato and Sekine(2013)}]{ShinzatoS13}
Keiji Shinzato and Satoshi Sekine. 2013.
\newblock \href {https://www.aclweb.org/anthology/I13-1190} {Unsupervised
  extraction of attributes and their values from product description}.
\newblock In \emph{Proceedings of the Sixth International Joint Conference on
  Natural Language Processing}, pages 1339--1347, Nagoya, Japan.

\bibitem[{Su et~al.(2020)Su, Zhu, Cao, Li, Lu, Wei, and Dai}]{SuZCLLWD20}
Weijie Su, Xizhou Zhu, Yue Cao, Bin Li, Lewei Lu, Furu Wei, and Jifeng Dai.
  2020.
\newblock \href {https://openreview.net/forum?id=SygXPaEYvH} {{VL-BERT:}
  pre-training of generic visual-linguistic representations}.
\newblock In \emph{8th International Conference on Learning Representations,
  {ICLR} 2020, Addis Ababa, Ethiopia, April 26-30, 2020}.

\bibitem[{Tan and Bansal(2019)}]{tan-bansal-2019-lxmert}
Hao Tan and Mohit Bansal. 2019.
\newblock \href {https://doi.org/10.18653/v1/D19-1514} {{LXMERT}: Learning
  cross-modality encoder representations from transformers}.
\newblock In \emph{Proceedings of the 2019 Conference on Empirical Methods in
  Natural Language Processing and the 9th International Joint Conference on
  Natural Language Processing (EMNLP-IJCNLP)}, pages 5100--5111, Hong Kong,
  China.

\bibitem[{Vaswani et~al.(2017)Vaswani, Shazeer, Parmar, Uszkoreit, Jones,
  Gomez, Kaiser, and Polosukhin}]{VaswaniSPUJGKP17}
Ashish Vaswani, Noam Shazeer, Niki Parmar, Jakob Uszkoreit, Llion Jones,
  Aidan~N. Gomez, Lukasz Kaiser, and Illia Polosukhin. 2017.
\newblock \href {http://papers.nips.cc/paper/7181-attention-is-all-you-need}
  {Attention is all you need}.
\newblock In \emph{Advances in Neural Information Processing Systems 30: Annual
  Conference on Neural Information Processing Systems 2017, 4-9 December 2017,
  Long Beach, CA, {USA}}, pages 5998--6008.

\bibitem[{Wu et~al.(2016)Wu, Schuster, Chen, Le, Norouzi, Macherey, Krikun,
  Cao, Gao, Macherey, Klingner, Shah, Johnson, Liu, Kaiser, Gouws, Kato, Kudo,
  Kazawa, Stevens, Kurian, Patil, Wang, Young, Smith, Riesa, Rudnick, Vinyals,
  Corrado, Hughes, and Dean}]{WuSCLNMKCGMKSJL16}
Yonghui Wu, Mike Schuster, Zhifeng Chen, Quoc~V. Le, Mohammad Norouzi, Wolfgang
  Macherey, Maxim Krikun, Yuan Cao, Qin Gao, Klaus Macherey, Jeff Klingner,
  Apurva Shah, Melvin Johnson, Xiaobing Liu, Lukasz Kaiser, Stephan Gouws,
  Yoshikiyo Kato, Taku Kudo, Hideto Kazawa, Keith Stevens, George Kurian,
  Nishant Patil, Wei Wang, Cliff Young, Jason Smith, Jason Riesa, Alex Rudnick,
  Oriol Vinyals, Greg Corrado, Macduff Hughes, and Jeffrey Dean. 2016.
\newblock \href {http://arxiv.org/abs/1609.08144} {Google's neural machine
  translation system: Bridging the gap between human and machine translation}.
\newblock \emph{CoRR}, abs/1609.08144.

\bibitem[{Xu et~al.(2019)Xu, Wang, Mao, Jiang, and Lan}]{XuWMJL19}
Huimin Xu, Wenting Wang, Xin Mao, Xinyu Jiang, and Man Lan. 2019.
\newblock \href {https://doi.org/10.18653/v1/P19-1514} {Scaling up open tagging
  from tens to thousands: Comprehension empowered attribute value extraction
  from product title}.
\newblock In \emph{Proceedings of the 57th Annual Meeting of the Association
  for Computational Linguistics}, pages 5214--5223, Florence, Italy.

\bibitem[{Yih et~al.(2015)Yih, Chang, He, and Gao}]{yih-etal-2015-semantic}
Wen-tau Yih, Ming-Wei Chang, Xiaodong He, and Jianfeng Gao. 2015.
\newblock \href {https://www.aclweb.org/anthology/P15-1128} {Semantic parsing
  via staged query graph generation: Question answering with knowledge base}.
\newblock In \emph{Proceedings of the 53rd Annual Meeting of the Association
  for Computational Linguistics and the 7th International Joint Conference on
  Natural Language Processing (Volume 1: Long Papers)}, pages 1321--1331,
  Beijing, China.

\bibitem[{Yu et~al.(2017)Yu, Yin, Hasan, dos Santos, Xiang, and
  Zhou}]{yu-etal-2017-improved}
Mo~Yu, Wenpeng Yin, Kazi~Saidul Hasan, Cicero dos Santos, Bing Xiang, and Bowen
  Zhou. 2017.
\newblock \href {https://www.aclweb.org/anthology/P17-1053} {Improved neural
  relation detection for knowledge base question answering}.
\newblock In \emph{Proceedings of the 55th Annual Meeting of the Association
  for Computational Linguistics (Volume 1: Long Papers)}, pages 571--581,
  Vancouver, Canada.

\bibitem[{Yu et~al.(2019)Yu, Yu, Cui, Tao, and Tian}]{Yu0CT019}
Zhou Yu, Jun Yu, Yuhao Cui, Dacheng Tao, and Qi~Tian. 2019.
\newblock \href
  {http://openaccess.thecvf.com/content\_CVPR\_2019/html/Yu\_Deep\_Modular\_Co-Attention\_Networks\_for\_Visual\_Question\_Answering\_CVPR\_2019\_paper.html}
  {Deep modular co-attention networks for visual question answering}.
\newblock In \emph{{IEEE} Conference on Computer Vision and Pattern
  Recognition, {CVPR} 2019, Long Beach, CA, USA, June 16-20, 2019}, pages
  6281--6290.

\bibitem[{Zheng et~al.(2018)Zheng, Mukherjee, Dong, and Li}]{ZhengMD018}
Guineng Zheng, Subhabrata Mukherjee, Xin~Luna Dong, and Feifei Li. 2018.
\newblock \href {https://doi.org/10.1145/3219819.3219839} {Opentag: Open
  attribute value extraction from product profiles}.
\newblock In \emph{Proceedings of the 24th {ACM} {SIGKDD} International
  Conference on Knowledge Discovery {\&} Data Mining, {KDD} 2018, London, UK,
  August 19-23, 2018}, pages 1049--1058.

\end{thebibliography}
\bibliographystyle{acl_natbib}

\end{CJK*}
\end{document}